\documentclass{article}
\usepackage{graphicx}
\usepackage{xcolor}

\PassOptionsToPackage{numbers, compress}{natbib}

\usepackage[preprint]{neurips_2025}

\usepackage[utf8]{inputenc} 
\usepackage[T1]{fontenc}    
\usepackage{hyperref}       
\usepackage{url}            
\usepackage{booktabs}       
\usepackage{amsfonts}       
\usepackage{amsmath}
\usepackage{nicefrac}       
\usepackage{microtype}      
\usepackage{xcolor}         

\title{Double Descent as a Lens for Sample Efficiency in Autoregressive vs. Discrete Diffusion Models}

\author{
\textnormal{Ahmad Fraij} \\
ETH Zurich \\
\texttt{afraij@ethz.ch} 
\and
\textnormal{Sam Dauncey} \\
ETH Zurich \\
\texttt{sdauncey@ethz.ch}
}

\begin{document}

\maketitle

\begin{abstract}
    Data scarcity drives the need for more sample-efficient large language models. In this work, we use the double descent phenomenon to holistically compare the sample efficiency of discrete diffusion and autoregressive models. We show that discrete diffusion models require larger capacity and more training epochs to escape their underparameterized regime and reach the interpolation threshold. In the strongly overparameterized regime, both models exhibit similar behavior, with neither exhibiting a pronounced second descent in test loss across a large range of model sizes. Overall, our results indicate that autoregressive models are more sample-efficient on small-scale datasets, while discrete diffusion models only become competitive when given sufficient capacity and compute. The code is available at: \textcolor{blue}{\url{https://github.com/Ahmad1732002/AR-vs-Diffusion-Sample-Efficiency}}.
\end{abstract}

\section{Introduction}
\label{sec:intro}
Recent progress in large language modeling has been driven primarily by scaling: training increasingly larger models on larger datasets to achieve substantial performance improvements \cite{kaplan2020scalinglawsneurallanguage,hoffmann2022trainingcomputeoptimallargelanguage}. This dependence on massive datasets implies that the demand for human-generated text is growing rapidly. However, such scaling is not sustainable. The stock of public human-generated text may be exhausted within this decade \cite{villalobos2024rundatalimitsllm}, highlighting a fundamental limitation for continued scaling.

One way to mitigate this data scarcity is to improve the sample efficiency of generative models—that is, to design or adopt architectures that achieve stronger performance with less data. By analyzing and comparing the sampling efficiency of different models, we can identify those that make better use of available data and reduce the dependence on huge datasets.

We investigate discrete diffusion models \cite{austin2023structureddenoisingdiffusionmodels}, whereby the model learns to patch a corrupted passage of text as a potentially more sample-efficient method of pretraining large language models compared to the more common next-token prediction objective of autoregressive models. We propose discrete diffusion models as a promising direction in the search for sample-efficient large-language model pretraining for two reasons:
(1) Prior work has shown that autoregressive models overfit to the direction in which facts are presented to them \cite{zhu_2024_physics_of_language_models, berglund_2024_reversal_curse} but that the masked-language modeling objective can mitigate this \cite{kitouni2024factorization}. (2) In the generative image task, despite being introduced in 2015 \cite{sohldickstein2015deepunsupervisedlearningusing}, diffusion models only broke through as state-of-the-art in 2020 \cite{ho2020denoisingdiffusionprobabilisticmodels}, with Ho et al.'s headline result on CelebA-HQ requiring $\approx 1000$ repetitions of each data sample. We hypothesize that this emergence of diffusion models on images could be the result of increasing compute budgets over time being applied to comparatively small fixed-size datasets such as CelebA-HQ, with $30,000$ samples.

We propose comparison of \emph{double descent curves} as a holistic way to measure the sample efficiency of these models in the data constrained setting. Double descent extends the classical bias–variance trade-off by revealing that as model capacity increases, test error does not monotonically decrease but instead follows a double descent curve: decreasing in the classical bias–variance regime, increasing near the interpolation threshold where models just interpolate the training data, and then decreasing again in the over-parameterized regime. 

It was first formalized in 2019 by Belkin et al. \cite{Belkin_2019}, who claimed that scaling the number of parameters in the model is the primary driver of double descent behavior. Subsequent work by Nakkiran et al. \cite{nakkiran2019deepdoubledescentbigger} generalized this perspective by revealing that double descent arises along multiple axes, including model size, training epochs, and dataset size. 

By studying double descent behavior through scaling model size, we gain a clearer understanding of how sample-efficient different models are when trained on limited data. A more sample-efficient model generalizes well with fewer parameters. In double descent terms, this means a more sample-efficient model maintains lower test loss in the underparameterized regime, achieves better optimal test performance, reaches the interpolation threshold at a larger model size, and experiences a faster decrease in test loss in the overparameterized regime.

Concurrent work by Prabhudesai et al. \cite{prabhudesai2025diffusionbeatsautoregressivedataconstrained} compares autoregressive (AR) and diffusion models in data-constrained regimes, analyzing validation loss across model sizes, compute budgets, and repeated data exposure. We contribute to this work by examining the behaviour of these models in the extremely overparameterized regime, and examining the  non-monotonic generalization behavior that arises past the interpolation threshold.

We leverage the double descent framework to provide a systematic comparison of autoregressive and discrete diffusion models on the extremely small-scale Shakespeare dataset. We empirically show that:

\begin{itemize}
    \item The double descent curve for discrete diffusion models has a broader underparameterised regime compared to autoregressive models.
    \item The double descent curve for diffusion models appears after a much larger number of epochs than autoregressive models (when equating the number of epochs in each model by training flops).
    \item Nonetheless, for our comprehensive sweep of model sizes and epoch counts for a small text dataset, the variational bound on the diffusion model's test bits-per-token (BPT) is higher than the BPT of the autoregressive model.
    \item For our setup, neither autoregressive nor diffusion models exhibit a pronounced second descent, with test loss remaining high even as the models are scaled past the interpolation thereshold, even with weight decay applied.
\end{itemize}


\section{Experimental Setup}
\label{sec:exp_setup}

\subsection{Model Training}

Autoregressive and Diffusion models are trained using an exact and an upper bound on the negative log-likelihood, respectively, allowing for their direct comparison. 

\subsubsection{Autoregressive Language Model}

Autoregressive models generate sequences by predicting each token conditioned on all preceding tokens. For a training sequence be \(x = (x_1, x_2, \dots, x_L)\), where \(L\) is the sequence length and each token \(x_t\) belongs to the vocabulary \(\mathcal{V}\). We train the model to minimize the bits-per-token (BPT) loss, which can be computed exactly via the autoregressive factorisation:
\[
\text{BPT}(\theta) = - \frac{1}{L} \log_2 p_\theta(x_1 \dots x_L) 
= - \frac{1}{L} \sum_{l=1}^{L} \log_2 p_\theta(x_l \mid x_{<l})
\]

\subsubsection{Discrete Diffusion Model}
 In discrete diffusion models, the forward corruption process of discrete data, such as text tokens, is done by applying a categorical transition matrix on the one-hot encoded text tokens \cite{austin2023structureddenoisingdiffusionmodels}. After a series of \(T\) forward diffusion steps the distribution of discrete tokens approaches a stationary distribution over \(K = |\mathcal{V}| + 1\) categories, where \(|\mathcal{V}|\) is the size of the original vocabulary and the extra category, added at index \(K-1\), corresponds to the absorbing \([MASK]\) token.

In our experiments we choose the transition matrix to use the same absorbing state as in Austin et al. \cite{austin2023structureddenoisingdiffusionmodels} meaning that each token is either unchanged or transitions to a \([MASK]\) state with probability $\beta_t = \frac{1}{T - t + 1}$, hence after \(T\) diffusion steps it approaches a non-uniform stationary distribution that has its mass concentrated on the \([MASK]\) token. Specifically, we use a forward process defined by:

\[
    q(x_{l}^{(t)} \vert x_{l}^{(t-1)} ) = 
    \begin{cases}
        \beta_t & \text{if }  x_{l}^{(t)} = [MASK] \\
        (1 - \beta_t) & \text{if } x_{l}^{(t)} = x_{l}^{(t-1)} \\
        0 & \text{otherwise}
    \end{cases}
\]

Whereby each token begins uncorrupted, ie $x_l =x_{l}^{(0)}$.

A transformer-based model is trained to denoise the model and predict the true categorical distribution over \(\mathcal{V}\) categories.

For discrete diffusion models, we have an upper bound on the negative log-likelihood, given by:
\begin{equation*}
\begin{gathered}
\text{BPT}(\theta) = -\log_2 p_\theta(x_1 \dots x_L) \\
\leq \frac{1}{\ln 2} \mathbb{E}_{q}\big[-\log p_\theta(x_{:}^{(0)} \mid x_{:}^{(1)})\big] 
+ \frac{1}{\ln 2} \sum_{t=2}^{T} \mathbb{E}_{q}\Big[\mathrm{KL}\big(q(x_{:}^{(t-1)} \mid x_{:}^{(t)}, x_{:}^{(0)}) \,\|\, p_\theta(x_{:}^{(t-1)} \mid x_{:}^{(t)})\big)\Big]
\end{gathered}
\end{equation*}

where \(x_i\) denotes the true value of token \(i\), and \(x_{:}^{(t)}\) represents the full sequence after corruption at diffusion timestep \(t\). We defer a complete derivation to  Austin et al. \cite{austin2023structureddenoisingdiffusionmodels}.  
The first term corresponds to the reconstruction cross-entropy at \(t=0\), while the second term sums KL divergences between the true posterior and the model predictions for \(t=2,\dots,T\). During training and validation, we do a Monte-Carlo estimate of the upper bound on the negative log-likelihood by sampling one $t$ value per sample per epoch. During validation, we use the same procedure per sample but average over 100 batches to obtain a low-variance estimate of the bound.

\subsection{Training Details / Hyperparameters}

To ensure a fair comparison of sample efficiency between the two generative models, we train both models using the same standard transformer architecture \cite{Radford2019LanguageMA}. The model consists of a stack of $L=12$ transformer decoder blocks. Each block applies multi-head causal self-attention with rotary positional embeddings (RoPE) \cite{su2023roformerenhancedtransformerrotary}, followed by a feed-forward network composed of two linear layers with GELU activation \cite{hendrycks2016GELU}. Residual connections and layer normalization are applied after both the attention and feed-forward layers. Nonetheless, the diffusion model requires two key architectural distinctions: 1) the self-attention is non causal (meaning that every token can attend to tokens succeeding it as well as preceding it) and (2) we use a multi-layer perceptron (MLP) to embed the timestep in order for the model to know how corrupted each text token is at each specific diffusion step.

In appendix~\ref{sec:appendix} We provide a full list of hyperparameters in Table~\ref{tab:hyperparams} and training resources used for the experiments.

\subsection{Dataset \& Preprocessing}
\label{sec:dataset}

We use the Tiny Shakespeare dataset consisting of roughly 40{,}000 lines extracted from various plays by Shakespeare. We split the dataset into 90\% training and 10\% validation. We tokenize the text using the GPT-2 Byte-Pair Encoding (BPE) tokenizer. This preprocessing yields approximately 302K training tokens and 36K validation tokens.

\section{Results}
\label{sec:Results}

\begin{figure}[h]  
    \centering
    \includegraphics[width=1.0\textwidth]{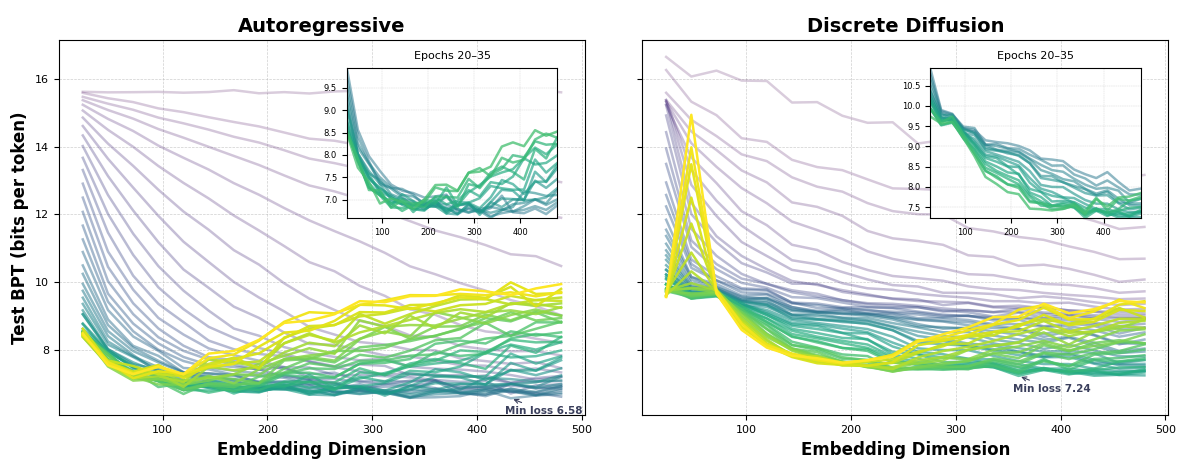}
    \includegraphics[width=1.0\textwidth]{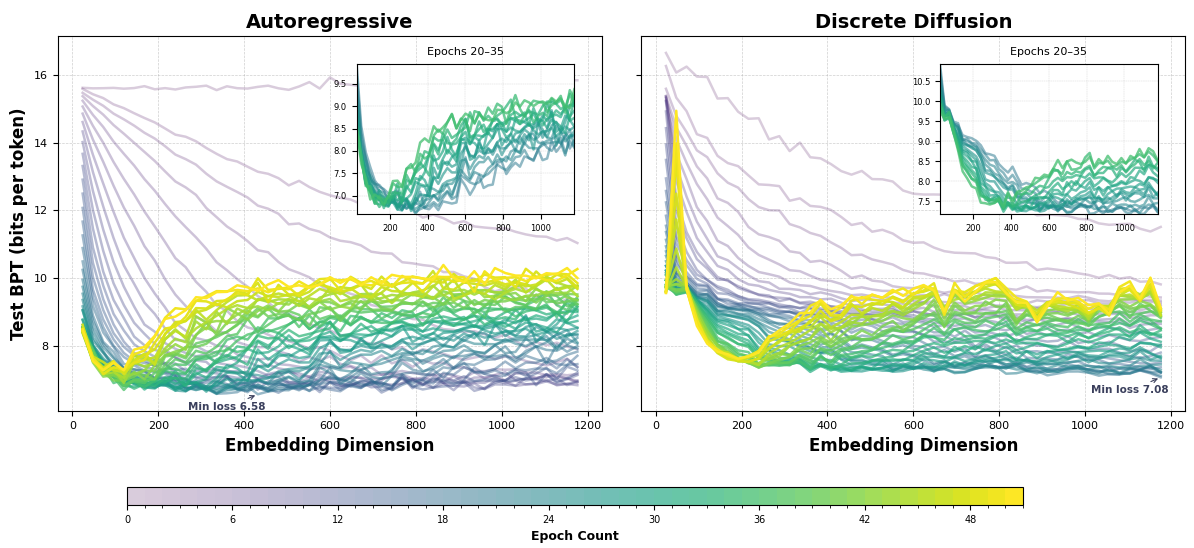}
    \vspace{1ex}  
    \caption{Test BPT (bits per token) as a function of embedding dimension for autoregressive (left) and discrete diffusion (right) models across training epochs. Curves are colored by epoch count. Insets zoom into epochs 20–35, where double descent first becomes visible (around epoch 21 for autoregressive and epoch 33 for discrete diffusion)}
    \vspace{1ex}  

    \label{fig:epoch_vs_error}
\end{figure}

We evaluate sample efficiency through the lens of double descent. Specifically, we examine how quickly models overfit the data as a function of embedding dimension and training epoch, how the interpolation threshold shifts with model size, and how the bits-per-token (BPT) loss evolves in the underparametrized and overparameterized regimes (Figure~\ref{fig:epoch_vs_error}).

\subsection{Performance Across Model Sizes and Epochs}
For the diffusion model, the interpolation threshold (where the test BPT loss peaks) consistently occurs at larger embedding dimensions around 600 compared to autoregressive models around 400. This shift indicates that diffusion requires greater model capacity to both reach its optimal BPT loss and to cross into the overparameterized regime, resulting in a broader underparameterized regime across all epochs.

By examining the zoomed-in plots, it is apparent that the double descent behavior, including the first rise in the BPT test errors, occurs only after 21 epochs in the autoregressive case, compared to 33 epochs in the discrete diffusion case.

Across all epochs and embedding dimensions, the autoregressive model achieves a lower optimal BPT error of 6.58 at epoch 18 with an embedding dimension of 432. In comparison, the discrete diffusion model reaches an optimal BPT error of 7.08 at epoch 31 with an embedding dimension of 1176.

These results shows clear differences in sample efficiency between the two architectures. The autoregressive model reaches its optimal BPT loss at smaller embedding dimensions and earlier epochs, demonstrating stronger generalization with fewer parameters and less training data. In contrast, the diffusion model requires larger embedding dimensions and more compute to achieve its optimum, reflecting lower immediate sample efficiency. However, its broader underparameterized regime indicates a more gradual utilization of model capacity, which can be advantageous when scaling up. This shows that autoregressive models are more immediately sample-efficient on this clearly labeled tiny Shakespeare data, and diffusion models offer a more scalable generalization and are less prone to overfitting as model capacity scales.

\subsection{Performance with Early Stopping}

\begin{figure}[h]  
    \centering
    \includegraphics[width=0.9\textwidth]{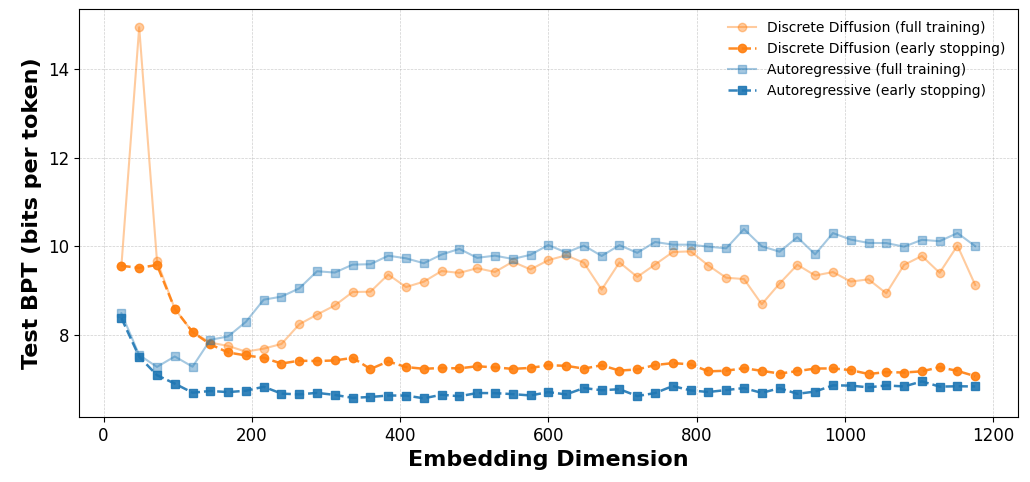}
    \vspace{1ex}  
    \caption{Comparison of double descent behaviours for autoregressive and discrete diffusion models with different embedding dimensions.Solid lines with lower opacity show the loss after 50 epochs, while dashed lines with higher opacity show the BPT loss with early stopping at the best epoch for each embedding dimension}

    \vspace{1ex}  

    \label{fig:model_size_vs_error}
\end{figure}

To further analyze the sample efficiency of both models, we examine their behavior under early stopping. Early stopping provides a direct measure of the best generalization performance attainable at each model size. Comparing these minima across architectures reveals which model reaches lower optimal BPT loss with fewer parameters, and is therefore more sample efficient.

When early stopping occurs (Figure \ref{fig:model_size_vs_error}), the autoregressive model achieves lower BPT loss values along all embedding dimensions compared to the discrete diffusion model, which shows it is clearly more sample efficient in that case. The double descent behavior is not observed for either models, as they are not overfitting the data, and the test error converges for both models with increasing embedding dimensions.

When both models are fully trained for 50 epochs, beyond their early-stopped optimal test loss values, the discrete diffusion model surpasses the autoregressive model at embedding sizes larger than 200, demonstrating better generalization when sufficient compute and model capacity is provided.

It is also clear that the interpolation threshold and the overparameterized regimes become more apparent for both models once they are fully trained on 50 epochs (Figure~\ref{fig:model_size_vs_error}). In these regimes, the BPT loss of both models remains high and does not exhibit a pronounced second descent.

\section{Limitations}
\label{sec:Limitations}
Our experimental results are limited by the fact that they were conducted only on the tiny-Shakespeare dataset. This dataset is highly structured and cleanly labeled, so the double descent behaviors observed may differ on less structured data with more label noise. Prior work has shown that adding label noise to small datasets can lead to more pronounced double descent\cite{dubost2021doubledescentoptimizationpattern}. Additionally, we equate epochs via flops (doing one forward/backward pass per sequence in the train dataset), however each forward/backward pass for the autoregressive model equates to predicting all the tokens in the sequence, whereas for a diffusion model only equates to predicting some subset of tokens in the sequence.

\section{Conclusion}
We investigate the sample efficiency of autoregressive and discrete diffusion models on the tiny-Shakespeare dataset. Our analysis shows that discrete diffusion models tolerate underparameterization over a broader range of model sizes. However, they require more training epochs to achieve competitive performance, and their optimal test bits-per-token (BPT) remains higher than that of autoregressive models. This indicates that, in practice, they are not strictly more sample-efficient on this small scale dataset. Furthermore, we report no second descent in our training setup, suggesting that the pronounced second descent that is seen in image classification \cite{Belkin_2019, nakkiran2019deepdoubledescentbigger} may not apply to generative language modelling. Future work could extend this analysis to larger datasets with varying levels of label or input noise, revealing how robust these architectures are to imperfect data.

\bibliography{references}

\newpage
\appendix

\section{Technical Appendices and Supplementary Material}
\label{sec:appendix}
\subsection{Hyperparameters}
\label{sec:hyperparams}
The Model depth and number of attention heads are chosen to have relatively small models that can be trained without huge compute requirements. The sequence length and vocabulary size are chosen to match the small Shakespeare dataset and its tokenization. Batch sizes are adjusted to fit GPU memory constraints, particularly for the diffusion model due to its per-step noise computations. Learning rates and optimizer are standard settings for stable transformer training. Diffusion-specific hyperparameters (number of steps $T$ and CE loss weight $\lambda_{CE}$) are chosen to ensure effective denoising without destabilizing training.

We base our code off of the \texttt{nano-gpt} \footnote{\url{https://github.com/karpathy/nanoGPT}} and \texttt{nanoDD} repositories \footnote{\url{https://github.com/flukeskywalker/nanoDD/tree/main}} respectively
\begin{table}[h]
\centering
\caption{Comparison of model and training hyperparameters for the trained autoregressive and discrete diffusion models}
\label{tab:hyperparams}

\begin{tabular}{lcc}
\toprule
\textbf{Hyperparameter} & \textbf{Autoregressive} & \textbf{Discrete Diffusion} \\
\midrule
Vocab size & 50257 & 50257 \\
Number of heads & 12 & 12 \\
Number of layers & 12 & 12 \\
Conditional embedding size & -- & 128 \\
Sequence length & 512 & 512 \\
Batch size & 12 & 8 \\
Learning rate & 6e-4 & 6e-4 \\
Weight decay & 0.1 & 0.1 \\
Optimizer & AdamW & AdamW \\
Diffusion steps ($T$) & -- & 1000 \\
CE loss weight ($\lambda_{CE}$) & -- & 0.05 \\
\bottomrule
\end{tabular}

\end{table}

For training autoregressive models, we sample from the train dataset without replacement, whereas for training diffusion models sample from the train dataset with replacement, ensuring that the same number of sequences are sampled per epoch.

\subsection{Training resources}

\label{sec:compute}
The training for the experiments was conducted using RTX\_3090 with 24GB GPUs in a cluster environment. One GPU was used to run a single experiment which takes from 1-6 hours to complete depending on the model size and number of epochs. In total, 10 experiments were being conducted in parallel to speed up the overall training. Hence, 10 GPUs were being used and a total of 240 GB of memory was needed to run the experiments.

\bibliographystyle{unsrtnat} 
\end{document}